\documentclass[letterpaper, 10 pt, conference]{ieeeconf}  

\IEEEoverridecommandlockouts                              

\usepackage{graphics} 
\usepackage{graphicx}
\usepackage{mathptmx} 
\usepackage{cite}
\usepackage{amsmath} 
\usepackage{amssymb}  
\usepackage{bm}      

\title{\LARGE \bf
Inverted Landing in a Small Aerial Robot via Deep Reinforcement Learning for Triggering and Control of Rotational Maneuvers
}

\author{Bryan Habas$^{1}$,
Jack W. Langelaan$^{2},$ \textit{IEEE, Member}, 
Bo Cheng$^{1}$, \textit{IEEE, Member}
\thanks{$^{1}$Biological and Robotic Intelligent Fluid Locomotion Lab, Department of Mechanical Engineering, The Pennsylvania State University, University Park, PA 16802, USA. Corresponding to B.C. {\tt\small buc10@psu.edu}}
\thanks{$^{2}$Air Vehicle Intelligence and Autonomy Lab, Department of Aerospace Engineering, The Pennsylvania State University, University Park, PA 16802, USA}%
}

\begin{document}

\maketitle
\thispagestyle{empty}
\pagestyle{empty}

\begin{abstract}

Inverted landing in a rapid and robust manner is a challenging feat for aerial robots, especially while depending entirely on onboard sensing and computation. In spite of this, this feat is routinely performed by biological fliers such as bats, flies, and bees. Our previous work has identified a direct causal connection between a series of onboard visual cues and kinematic actions that allow for reliable execution of this challenging aerobatic maneuver in small aerial robots. In this work, we utilized Deep Reinforcement Learning and a physics-based simulation to obtain a general, optimal control policy for robust inverted landing starting from any arbitrary approach condition. This optimized control policy provides a computationally-efficient mapping from the system's emulated observational space to its motor command action space, including both triggering and control of rotational maneuvers. This was accomplished by training the system over a large range of approach flight velocities that varied with magnitude and direction. 

Next, we performed a sim-to-real transfer and experimental validation of the learned policy via domain randomization, by varying the robot's inertial parameters in the simulation.  Through experimental trials, we identified several dominant factors which greatly improved landing robustness and the primary mechanisms that determined inverted landing success. We expect the reinforcement learning framework developed in this study can be generalized to solve more challenging tasks, such as utilizing noisy onboard sensory data, landing on surfaces of various orientations, or landing on dynamically-moving surfaces.

\end{abstract}

\section{INTRODUCTION}

\begin{figure}[tp]
    \centering
    \includegraphics[width=\columnwidth]{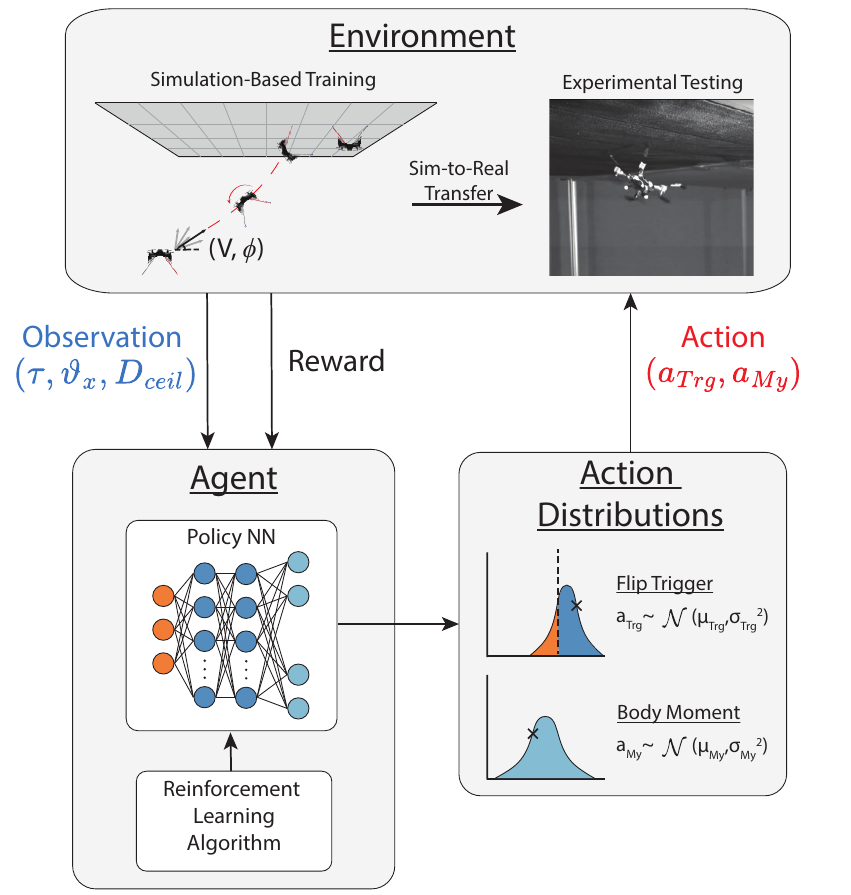}
    \caption{Reinforcement Learning diagram indicating the simulation-based training and experimental testing environments. As well, the flip triggering and motor command action distributions output by the policy network inside the agent.}
    \label{fig:Introduction_Fig}
\end{figure}

Landing on inverted surfaces is an aerial feat commonly exercised by bats, flies and bees \cite{bergou2015falling,liu2019flies,srinivasan2000honeybees}. Although aerial robots with more sophisticated mechanical designs are  capable of slow inverted landing (e.g., \cite{yu2020perching}); it remains a challenging problem to achieve rapid, robust inverted landing in small aerial robots with limited sensing and force vectoring capabilities. In fact, there lacks a systematic framework to develop a general control strategy that efficiently maps onboard sensor measurements to timely control actions for robustly triggering and controlling inverted landing under various conditions (e.g., different approaching velocities and angles). An aerial robot's ability to robustly generate inverted landings will greatly improve its operation capabilities for tasks like surveying unfamiliar terrain, solar charging of batteries, or perching on irregular surfaces\cite{mishra2020drone,kim2018drone,restas2015drone}.

Previous work by the authors, built upon bio-inspiration\cite{liu2019flies}, has investigated the landing strategies and landing gear designs for  robust inverted landing performance in small quadrotors \cite{habas2022optimal,liu2020bio}. However, their work only applies to a set of discrete and individually learned surface-approach conditions and cannot be generalized efficiently to a universal landing strategy applicable to arbitrary conditions. Another work by Mao and his coworkers \cite{mao2022robust}, have developed quadrotor systems that can land on inclined surfaces via onboard state estimation and trajectory planning. However, their methods depend on the computationally expensive process of generating and tracking trajectories. Additionally, work by Yu and their coworkers utilize force vectoring by rotational propellers which can generate inverted thrust to re-orient the quadrotor for a gentle inverted landing \cite{yu2020perching}.

The objective of this study is to develop a computationally-efficient, generalized control policy for  inverted landing. The control policy should utilize a minimal set of sensory cues, and  enable small aerial robots to land autonomously over a large range of flight conditions. For this purpose, we utilized a Deep Reinforcement Learning (Deep RL) algorithm to find an optimal policy that maps sensory cues obtainable from onboard measurements (e.g., time-to-contact, optical flow, and ceiling distance), to a set of triggering and control actions for dynamic landing maneuvers \cite{habas2022optimal}.

The sensory cues used in the control policy were inspired from those observed in animal perching behaviors. The first sensory cue used is a Time-to-Contact term, often represented as $\tau$, which encodes a system's time to contact with a surface (assuming a constant velocity and heading). This value has been shown to often be used in nature in connection to Tau-theory and has been well established as an effective action-perception cue for planning, collision avoidance, and landing in biological systems \cite{lee1993visual,lee1998guiding,yang2016bio}. $\tau$ can also be understood as the inverse of the Relative Retinal Expansion Velocity ($RREV$); which has been shown to be visually estimated by flying insects like flies\cite{liu2019flies}\cite{wagner1982flow}. The second sensory cue used is a visual observable term ($\vartheta_x$) derived from the fore/aft optical flow and encodes information about the system's translational velocity relative to a surface. Both of these terms can be accurately estimated through a single monocular camera sensor \cite{chirarattananon2018direct,horn2009hierarchical}. Lastly, the distance to the ceiling was used as the third sensory cue ($D_{ceil}$), which could be be calculated either through a laser distance sensor or with further estimation by fusion of the system's onboard accelerometer and time-derivative of $\tau$ \cite{dougherty2014laser,van2012visual,van2014monocular}. For the purpose of focusing on the landing strategy development, the above sensory values were emulated using external motion tracking data, as shown in the author's previous work \cite{habas2022optimal}, in lieu of a physical sensor or image processing algorithm. Follow-up work will incorporate these values with true onboard sensors and provide fully onboard landing capabilities.

In this work, a generalized inverted landing control policy was learned using the Soft Actor-Critic Deep Reinforcement Learning algorithm in a simulated environment and Sim-to-Real transfer techniques were used to transfer the learned policy to a physical quadrotor (Fig. \ref{fig:Introduction_Fig}). This control policy generated the required triggering timing for the rotational landing (or flip) maneuver  as well as the motor command for its magnitude over a varied array of approach velocities and flight angles. We also discuss the limitations of our experimental results due to the Sim-to-Real transfer and possible future work to improve landing robustness.

The rest of the paper is organized as follows. Section II provides a description of our methodology and Deep Reinforcement Learning algorithm implementation for generating an optimized landing policy. Section III details the learned strategy results acquired via simulation and our policy implementation in an experimental setting. Finally, Section IV concludes the study and provides directions for future work.

\section{METHODOLOGY}

In the authors' recent work \cite{habas2022optimal}, policy gradient learning was applied individually to an array of discrete approach conditions, from which a corresponding array of optimal state-action pairs were obtained for achieving reliable four-leg inverted landing. However, due to the limitations of working with a non-linear and discrete policy inside a continuous observation space, these results were insufficient to form a general inverted policy \cite{habas2022optimal}. Here, we instead trained a generalized landing policy using a neural network through Deep Reinforcement Learning (Deep RL), which is effective in a continuous observation and action space.

\subsection{Reinforcement Learning Background}

Reinforcement learning (RL) is a framework that makes use of machine learning to solve problems which can be broken down and defined as a Markov Decision Process (MDP). To build our framework to be compatible with future noisy-onboard sensors, we define our system as a finite-horizon, discounted, Partially Observable Markov Decision Process (POMDP). Here, the system dynamics are determined by an underlying MDP, where the agent cannot observe the state directly but only infer information about the state through noisy observations. At each timestep $t$, of this POMDP,  the agent makes an observation $\mathbf{o}_t \in O$ about the underlying state $\mathbf{s}_t \in S$, selects an action $\mathbf{a_t} \in A$, and receives a reward $r(\mathbf{s}_t,\mathbf{a}_t)$. At this point the system transitions into state $\mathbf{s}_{t+1}$ where the next state and observation are stochastically determined by the environment. In order to maximize the agent's performance, it must find a policy distribution $\pi_{\theta}(\mathbf{a}_t|\mathbf{o}_t)$, parameterized by $\boldsymbol{\theta}$, that produces trajectories where each observation is positively correlated with a high return $R$, and avoid trajectories with low return values. Where the return,  

\begin{equation}
    \label{Eq:R_t}
    R(s_t,a_t)=\sum_{k=0}^{T} \gamma^{k} r_{t+k+1},
\end{equation}

\noindent is the sum of discounted future rewards from the current timestep and the discount factor $\gamma \in [0,1)$ determines the extent the agent cares about rewards in the distant future.

Due to the purely episodic nature of our POMDP and slow speed of simulation, the off-policy Soft Actor-Critic (SAC) algorithm was chosen for its sample efficiency and excellent convergence properties \cite{haarnoja2018soft}. Additionally, SAC benefits from a modified objective function,
\begin{equation}
    \label{Eq:J_theta}
    J(\theta)=
    \mathbb{E}_{s_t \sim \rho^\pi,a_t \sim \pi_\theta}
    \left[R\left(s_{t}, a_{t}\right)+\beta H\left(\pi_{\theta}\left(s_{t}\right)\right)\right],
\end{equation}

\noindent which includes an entropy regularization term that encourages exploration while still maximizing the total return \cite{williams1991function}. By adding this entropy regularization term directly to the objective function, the policy is forced to be as non-deterministic as possible while also achieving as much reward as possible. The benefit of this is that stochastic policies have been shown to be the optimal policy behavior in control problems that infer state information like those from noisy sensor readings\cite{todorov2008general,toussaint2009robot}.

In standard on-policy algorithms such as Advantage Actor-Critic (A2C) \cite{mnih2016asynchronous} and Proximal Policy Optimization (PPO) \cite{schulman2017proximal}, exploration is often biased towards improving a single deterministic strategy and refining it to maximize the total reward. However, the detriment of this strategy is that the observation space can often be under-explored and other possible actions will be ignored in favor of optimizing the current policy and being trapped in a local optima. The benefit of combining Maximum Entropy RL and off-policy learning is a very efficient exploration strategy where the agent tries to reduce its own uncertainty about the environment through controlled exploration and gain more information about the environment than simply searching for a good policy would allow. This has shown to perform better than several other Deep RL algorithms due to an improved performance in finding a global optima \cite{haarnoja2018soft}.

\subsection{Reinforcement Learning Problem Formulation}

In this work we trained a control policy in a stochastic environment to solve an inverted landing problem for a quadrotor robot. The quadrotor was placed in the environment and set to follow a collision trajectory toward a ceiling object with a specified approach velocity and angle ($V$,$\phi$) relative to the horizon; where an angle of $0^{\circ}$ is parallel to the horizon and an angle of $90^{\circ}$ is purely vertical flight. Here, the goal of the Deep RL agent was to learn an optimal policy $\pi^{*}_{\theta}(\mathbf{a}_t|\mathbf{o}_t)$ that maximizes the total reward for an episode. At each timestep, the quadrotor made an observation about the environment through its emulated sensors (state information given by the simulation or external positioning system which mimic onboard sensory values \cite{habas2022optimal}), performed an action according the policy $\pi_{\theta}(\mathbf{a}_t|\mathbf{o}_t)$, and received a reward. 

\begin{figure}[t]
    \centering
    \includegraphics[width=\columnwidth]{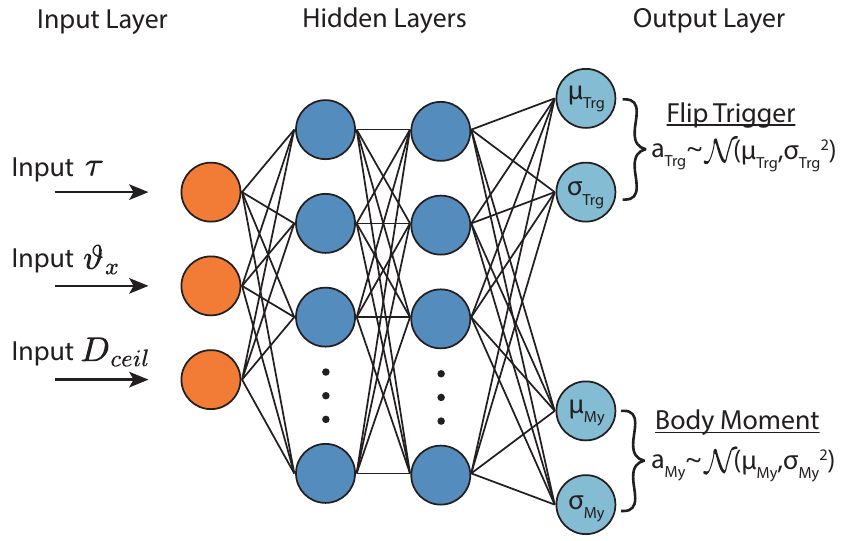}
    \caption{Event triggering policy network with two action heads that parameterize their respective Gaussian distributions. Each distribution is sampled at each timestep, where if the flip trigger action ($a_{Trg}$) is greater than a defined threshold, the motor command for the flip action ($a_{My}$) is applied.}
    \label{fig:Policy_Network}
\end{figure}

The observation space for this problem is similar to the augmented optical-flow space shown by Habas et al.\cite{habas2022optimal}, which includes the time-to-contact term ($\tau$), a scaled velocity term ($\vartheta_x$) related to translational optical flow \cite{chirarattananon2018direct}, and the distance to the ceiling ($D_{ceil}$). For this work, these terms were emulated with data received from a Vicon motion capture system to focus on the control strategy and implementation; with future work directly incorporating these terms with onboard sensors.

Due to the purely episodic nature of our problem \cite{habas2022optimal}, where only the triggering observations and final state of the system matter, all states but the triggering timestep, receive a reward of zero and all rewards from the final state are attributed to the action and observations that triggered the flip maneuver. However, by adjusting a discount factor to have a value of $\gamma = 0.999$ the reward received at the triggering timestep is discounted and back-propagated to the observations in the enacted flight trajectory which brought the system into the flip state, this process allows for proper learning of when to trigger the flip maneuver.

The policy network consisted of two separate actions heads as shown in Figure \ref{fig:Policy_Network}. The first being an event-triggering action head \cite{baumann2018deep,vamvoudakis2018model} that triggers the flip maneuver based on the system's current observation. As well as another action head that determines the rotation magnitude applied by the quadrotor's front rotors at the time of flip-triggering.

\begin{figure}[t]
    \centering
    \includegraphics[width=\columnwidth]{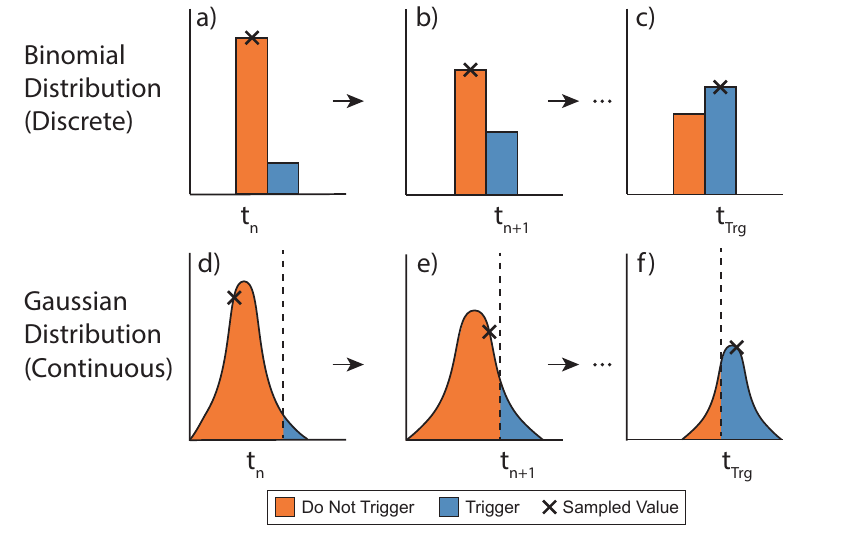}
    \caption{(a-c) At each timestep, a discrete distribution is updated and sampled until a binary value triggers a predefined event. (d-f) At each timestep, a Gaussian policy distribution is updated and sampled until the sampled value is greater than a defined threshold. At which point, the predefined event is triggered. }
    \label{fig:Flip_Trigger}
\end{figure}

To ensure triggering of the flip maneuver while in an optimal state, an action value ($a_{Trg}$) is sampled from a distribution parameterized by the flip triggering action-head during each timestep. This value is then used to trigger the flip maneuver depending on its Boolean $True$ or $False$ value. By doing so, the policy can be optimized to output a distribution where triggering the flip maneuver has a low probability of being sampled in all states except those with the most likely inverted landing success. 

In many cases this distribution is defined as a binomial (discrete) distribution where the probability of the action happening depends on the output of the policy network (Figure \ref{fig:Flip_Trigger}a-c). However, due to limitations defined in the SAC algorithm, a fully continuous action distribution is required. To continue work with this same event-triggering framework, we combined a parameterized Gaussian distribution with an arbitrary and trivial threshold value to approximate the behavior of the binomial distribution (Figure \ref{fig:Flip_Trigger}d-f). Here, a value was similarly sampled at each timestep from the distribution output by the policy network, $a_{Trg} \sim \mathcal{N}\left(\mu_{Trg}, \sigma_{Trg}^2\right)$, and if the value was larger than the set threshold the flip maneuver was triggered. Similarly, the distribution was updated with its corresponding observations such that unfavorable flip states likely correlated with sampled values less than the threshold (Figure \ref{fig:Flip_Trigger}d,e) and optimal flip states sample values above the threshold (Figure \ref{fig:Flip_Trigger}f).

Finally, at the moment of flip triggering, the body-moment motor command was sampled in a similar way as the flip trigger timing from its respective action distribution $a_{M y} \sim \mathcal{N}\left(\mu_{M y}, \sigma_{M y}^2\right)$. The sampled value was then compressed into the $[-1,1]$ range through the hyperbolic-tangent function and then re-scaled to the workable value range $My \in [0.0,8.0]$ N*mm. The robot then continued applying this moment until the system rotated past $90^{\circ}$ where the motors were then turned off.

\subsection{Reward Function Design}

In reinforcement learning, proper design of the reward function is paramount to learning success, speed of convergence, and quality of the policy learned by the agent. To ensure a quick and reliable convergence in our problem, 
the reward function was designed on the basis of curriculum learning, where each learned trait in the inverted landing process expands the total reward the system can achieve in progressive stages and guide the system in learning a more complex behavior. These traits and their respective reward functions were characterized as to: minimize the distance to the ceiling as in (\ref{Eq:r_d}); trigger the flip maneuver before colliding with the surface like (\ref{Eq:r_tau}); optimize the impact angle so the fore-legs contact first represented in (\ref{Eq:r_theta}); refine the above conditions to ensure reliable inverted landing is achieved as in (\ref{Eq:r_legs}).

Equations (\ref{Eq:r_d}) and (\ref{Eq:r_tau}) include the constants ($c_0,c_1$) which are used to normalize their reward functions as well as modify the width of their clipped regions. By adjusting these terms the reward functions will encourage the system to converge onto a range of usable values without over-fitting to a specific value. This can be seen specifically in (\ref{Eq:r_tau}), which encourages the flip to occur within the optimal triggering range $\tau_{trg} \in [0.15,0.25]$ without dictating when to execute the maneuver \cite{habas2022optimal}.

Equation (\ref{Eq:r_legs}) is also modified by a penalty factor of $r_{legs} \leftarrow r_{legs}/3$, whenever there is body or propeller contact; this is used to discourage any final policy that could lead to bodily damage of the quadrotor.

\begin{equation}
    \label{Eq:r_d}
    r_{d} = clip\left(\frac{1}{|d_{min}|},0,c_0\right) \cdot \frac{1}{c_0},
\end{equation}

\begin{equation}
    \label{Eq:r_tau}
    r_{\tau} = clip\left(\frac{1}{|\tau_{trg}-0.2|},0,c_1\right) \cdot \frac{1}{c_1},
\end{equation}

\begin{equation}
    \label{Eq:r_theta}
    r_{\theta} = 
            \begin{cases} 
            \frac{|\theta_{impact}|}{120^{\circ}} & 0^{\circ} \le |\theta_{impact}| < 120^{\circ} \\
            1.0 & 120^{\circ} \le |\theta_{impact}| \le 180^{\circ} 
           
            \end{cases},
\end{equation}

\begin{equation}
    \label{Eq:r_legs}
    r_{legs} = 
            \begin{cases} 
            1.0 & N_{legs} = 3 \ || \ 4 \\
            0.5 & N_{legs} = 1 \ || \ 2 \\
            0 & N_{legs} = 0 \\
            
            \end{cases}.
\end{equation}

Therefore, the total reward was weighted and combined, $r = 0.05\cdot r_{d} + 0.1\cdot r_{\tau} + 0.2\cdot r_{\theta} + 0.65\cdot r_{legs}$, to provide the total reward for the entire episode and attributed to the set of observations at the moment of flip triggering. The reason for this is that the remaining landing behavior was entirely defined from the observation-action pair taken at that timestep and any additional actions would confuse the agent and reduce learning performance.

\subsection{Simulation Setup and Sim-to-Real Transfer}

To accurately model the quadrotor system via simulation, several improvements over our previous work \cite{habas2022optimal} have been made in the system identification and modeling process \cite{habas2023TRO}. These included an improved estimation of the body's inertia matrix by use of a bifillar (two-wire) pendulum \cite{jardin2009optimized,de2011modeling}. As well as utilizing a tachometer to estimate the motor speed-up/slow-down characteristics and model the motor thrusts as a first-order dynamic system. These simulation improvements greatly improved the policy transfer behavior between simulation and experimental tests and allowed for a direct transfer between the two. 

For the Sim-to-Real transfer, Domain Randomization was utilized \cite{tobin2017domain}, i.e., before each rollout the system's mass and moment of inertia about the flip-axis were varied by sampling values from a Gaussian distribution with a standard deviation of $\sigma_m = 0.5$ g about the weighed mass and a standard deviation of $\sigma_{I} = 1.5\cdot10^{-6}$ kg*m\textsuperscript{2} about the estimated pitch-axis inertia. By doing so, a more robust policy can be found, which has been shown to improve experimental performance beyond trials without this method \cite{tobin2017domain,molchanov2019sim}.

The simulated training environment was configured such that the quadrotor would be launched at a randomly sampled velocity ($V \sim \mathcal{U}[1.5,3.5]$ m/s) and flight angle ($\phi \sim \mathcal{U}[30^{\circ},90^{\circ}]$) for each episode. By doing so, the system learned a generalized policy over the flight ranges that it was subjected to and optimized the policy for consistent inverted landing performance.

\begin{figure}[tp]
    \centering
    \includegraphics[width=\columnwidth]{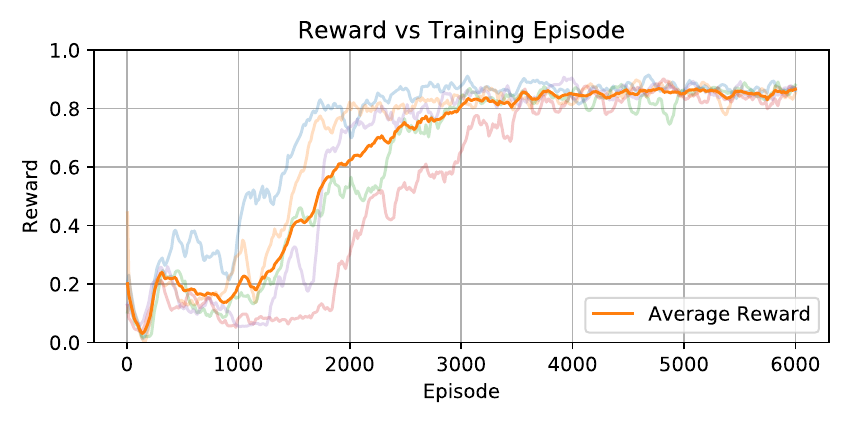}
    \caption{Reward convergence plot showing consistent convergence to a generalized inverted landing policy within 3000 episodes/rollouts. The plot shows the convergence history for several training sessions and their average convergence behavior.}
    \label{fig:Reward_Plot}
\end{figure}

\section{RESULTS AND DISCUSSION}

\subsection{Simulation-Based Training and Results}

Training was completed in the Gazebo simulation environment using the SAC algorithm provided by the StableBaselines3 RL package \cite{raffin2021stable}. In each training scenario, policy convergence  typically occurred within 3000 episodes/rollouts (Figure \ref{fig:Reward_Plot}). Compared to the previous work by Habas et al., which  used the parameter-exploring policy gradient method applied individually to each set of flight conditions \cite{habas2022optimal}, the Deep RL method required approximately $14\%$ of the number of rollouts to cover similar velocity and angle combinations \cite{habas2023TRO}. In addition, the output is a continuous policy which can handle untested observation values within the training range. Due to the nature of SAC maximizing both the reward and entropy of the system, there exists a large variance throughout the learning curve (Figure \ref{fig:Reward_Plot}). This behavior can be understood as the system exploring various state-action combinations to find the optimal policy which caused the curve to vary wildly at times. However, it was observed that this behavior still resulted in a robust convergence towards an optimal policy and consistently improved the reward across the learning time frame.

\begin{figure}[t]
    \centering
    \includegraphics[width=1.0\columnwidth]{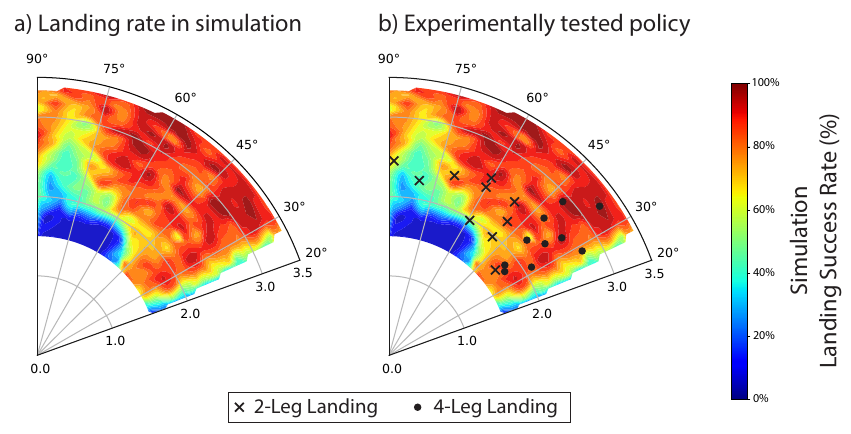}
    \caption{(a) Plot showing the landing success rate of the trained policy over the simulated flight conditions, $V \in [1.5 - 3.5]$ m/s and $\phi \in [25^{\circ} - 90^{\circ}]$. (b) Markers indicating the success of the simulation-based policy transferred to experimental trials.}
    \label{fig:Experiment_Landing_Rates}
\end{figure}

By sweeping over a large set of velocity and angle combinations and testing each with the optimized policy, the effective landing rate for each $(V,\phi)$ combination can be found as shown in Figure \ref{fig:Experiment_Landing_Rates}a. In this polar plot, the radial lines show the total flight speed and the angular lines dictate the flight angle as the quadrotor would approach the ceiling. Each location on the plot shows the corresponding landing success rate for the given flight conditions out of 30 attempts. 

Across all landing conditions, the consistent and optimal landing strategy observed was to maximize the rotational momentum of the body about the fore-leg contact point. In much the same way as a pendulum swinging, by maximizing this rotational momentum the robot would ensure that the hind-legs would swing up and finish the inverted landing maneuver. Therefore, by maximizing the transfer from the system's flight (or translational) momentum to rotational momentum after contact, the landing robustness and can be greatly improved. This can be accomplished either by controlling the body orientation at impact, leg geometry, or the flight angle as shown in \ref{fig:Experiment_Landing_Rates}a.

Throughout our policy validation trials in the simulation, it can be seen that in cases with a primarily vertical flight, a large flight velocity or (translational momentum) was needed to achieve reliable inverted landing success. However, a large increase in robustness can be observed in flight trajectories with a lower flight angle.
Here, this can be understood as the leg geometry and flight conditions creating a more efficient transfer of flight momentum into rotational momentum about the contact point. By observing this, it can be seen that this behavior resulted in both more reliable landing behaviors as well as increase landing performance at low speeds. 

Across all conditions, maximizing flight momentum when close to contact greatly improved landing robustness. By maximizing this overall momentum, landing success greatly improved although the impact forces will increase.  It can also be noted that failed landings in the regions shown were most often cases in which there was successful two-leg contact but insufficient body-swing to bring the remaining legs up to the ceiling surface. 

During the learning process, the value of the rotational moment provided by the motors was learned to be maximized under all conditions. Which indicates that a higher value will directly benefit the landing success and by reducing this value landing robustness will decrease. This can be understood as reducing the total rotational momentum of the system, which there is no benefit to do so. To extract a clear insight into the impact of the rotational moment, the Deep RL problem could be further constrained to reduce the energy expended in the landing process which would lead to an optimized policy that maximizes both the landing robustness and energy optimized landing strategy under all conditions.

\subsection{Experimental Validation and Implementation of Simulation Learned Policy}

Inverted landing experiments were performed with a nano-sized quadrotor equipped with four upgraded brushed DC motors. For these experiments, the Bitcraze Crazyflie 2.1 was chosen for its small mass, cheap and easily replaceable parts, and open-source firmware which together make the system a good choice for experimental testing and collision resilience. The legs were designed using the Narrow-Long configuration \cite{habas2022optimal} and equipped with a VELCRO\textsuperscript{\texttrademark} tip to attach to the ceiling. Communication to the robot was done through the Crazyswarm package \cite{preiss2017crazyswarm} which connects Robot Operating System (ROS) messages and Bitcraze's Crazyflie Real Time Protocol (CRTP) which allows for real-time data logging and command streaming. State data was streamed from a Vicon motion capture system at a frequency of 100 Hz and was used purely for sensor emulation to serve as a proxy for similar sensor values, as defined in previous work \cite{habas2022optimal}, and recorded with a mounted camera or optical flow sensor.

Experimental tests were completed by uploading the simulation-trained policy network to the Crazyflie, which would execute the policy actions at each timestep based on the current observation values. Sim-to-Real transfer was done in a one-shot manner without any further training in the experimental setup due to the collision nature of our problem and number of episodes needed to significantly continue training. In each experimental test, a desired velocity and angle would be given and a flight trajectory would be generated to accelerate the robot up to speed and on a collision course with the ceiling surface. At each timestep in the flight trajectory, the system would consistently check the policy output to decide if the flip maneuver should be triggered based on the current observations. Once triggered and the flip executed, the number of leg contacts and whether body contact was made was then recorded.

The resulting landing performance can be seen in Figure \ref{fig:Experiment_Landing_Rates}b and by video at ({\fontfamily{pcr}\selectfont https://youtu.be/fmDbaHJ15O8}), which shows successful landings  with high robustness in a similar region as the simulation. Here, low flight angles result in reliable four-leg landings (indicated by black dots) due to their efficient transfer of flight momentum to rotational momentum which the Narrow-Long leg design provided. Even in conditions where landing success rate was lower in the simulation, experimentally the quadrotor still managed to achieve a two-leg contact without body or propeller collisions. The two data sets overall match well, where regions in the simulation with the highest success rate correlate to similar behavior in the experimental setup. In summary, regions in the simulation which resulted in a lower success rate typically result in two-leg landings, and the regions in the simulation which resulted in a high success rate typically resulted in four-leg landings. Due to acceleration limits of the quadrotor, only a sub-region of the trained velocity and flight angle conditions could be tested.

\begin{figure}[t]
    \centering
    \includegraphics[width=0.9\columnwidth]{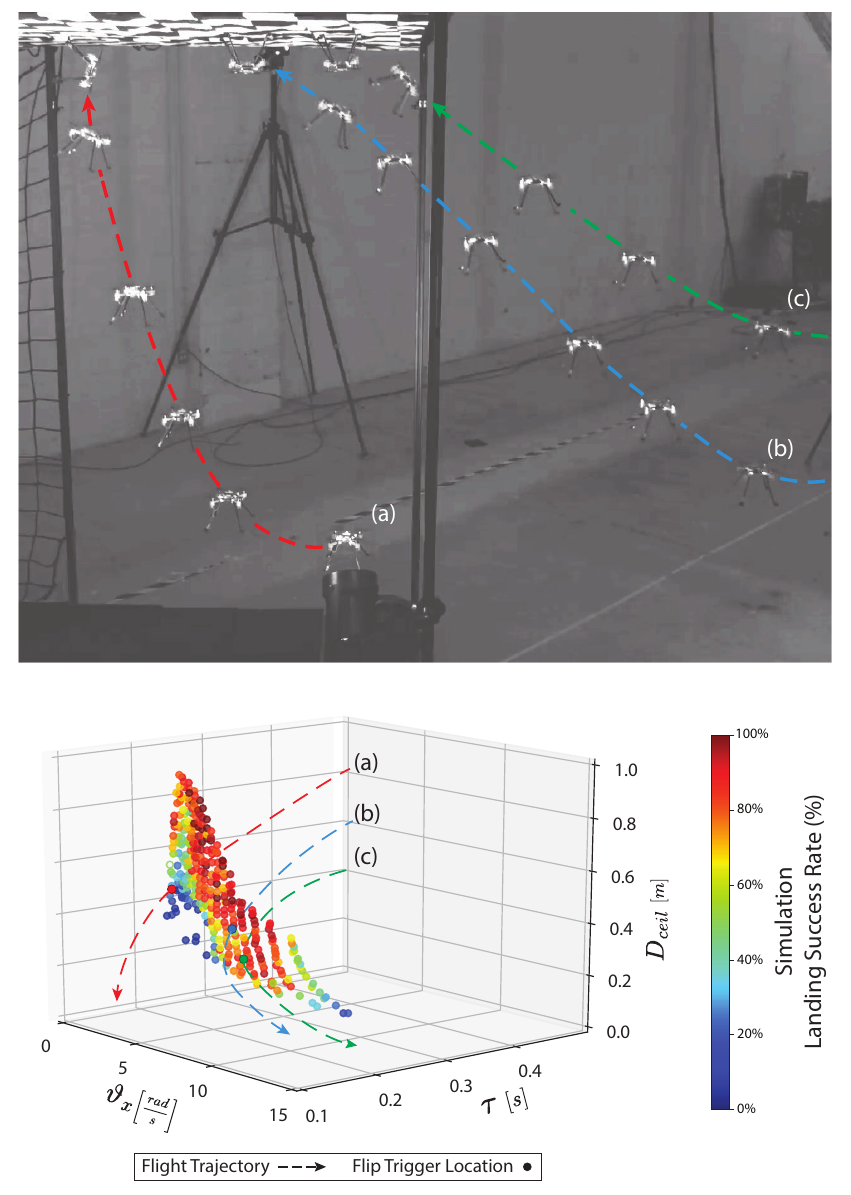}
    \caption{Example flight trajectories for experimental inverted landing tests (Top) and their corresponding trajectories through observation space (Bottom). (a) Near vertical flight trajectory ($2.25$ m/s, $70^\circ$) with two-leg landing. (b) Angled flight trajectory ($2.75$ m/s, $40^\circ$) with successful four-leg landing. (c) Angled flight trajectory ($2.50$ m/s, $40^\circ$) with successful four-leg landing.}
    \label{fig:DeepRL_Policy_Region}
\end{figure}

By mapping the collected landing rate data over the tested velocity conditions, the augmented triggering sensory-space can be mapped with regard to the landing success rate; resulting in a policy region (Fig \ref{fig:DeepRL_Policy_Region}) reminiscent of previous work\cite{habas2022optimal}. As well, a series of experimentally tested flight trajectories and inverted landings can be seen; including their trajectories through the augmented observation space and policy triggering region.

\section{CONCLUSIONS AND FUTURE WORK}

In this work we used a physics-based simulation and a Deep Reinforcement Learning algorithm to find a generalized inverted landing policy for a small aerial robot that is effective over a large array of initial approach conditions. This policy was then directly transferred to a physical quadrotor which performed robust inverted landings similar to those achieved in simulation. This work built upon previous work by Habas et al., which not only yielded a generalized policy but also improved the data efficiency in policy learning. We expect future work on using noisy real-time onboard sensory measurements for inverted landing will benefit from this deep learning method and control policy. 

This framework of determining the necessary flip action from sensor observations, as well as the knowledge of desirable velocity conditions that exhibit the most robustness, will allow for efficient use of future path-planning and trajectory optimization problems which target reaching the mapped policy region. By doing so, a complete landing sequence can be performed to maximize landing robustness, including the flight path planning into the policy region, and collision avoidance maneuvers when landing is not feasible. Additionally, the framework developed can be extended to learn landing strategies over a large range of surface orientations.  

Experimental and simulation results both confirmed that flight velocity was the dominant factor in landing success and that low flight angles exhibited the greatest robustness due to advantages of the leg geometry and efficient transfer of translational momentum to rotational momentum about the contact point, which allows for efficient swing up of the hind-legs after fore-leg contact. Interestingly, similar behaviors are also observed in fliers \cite{liu2019flies}. 

\addtolength{\textheight}{-12cm}   


\section*{ACKNOWLEDGMENT}

This research was supported by the National Science
Foundation (IIS-1815519 and CMMI-1554429) awarded to B.C. and supported by the Department of Defense (DoD) through the National Defense Science \& Engineering Graduate (NDSEG) Fellowship Program awarded to B.H.

\bibliography{refs.bib}
\bibliographystyle{IEEEtran}

\end{document}